\documentclass[letterpaper, 10 pt, conference]{ieeeconf}
\IEEEoverridecommandlockouts
\usepackage[noadjust]{cite}
\usepackage{subfig}
\usepackage{amsmath,amssymb,amsfonts}
\usepackage{graphicx}
\usepackage[table]{xcolor}
\usepackage{caption}

\title{\LARGE \bf Development of Low-Cost and Bidirectional Syringe Pumps for Soft Robotics Applications}

\author{
    Krishamsu Subedi Chhetri, Aryan Mayor, Elise Corbin, Logan Walker, John Rieffel\\ 
   {\textit{\tt\small Evolutionary Robotics Laboratory}} \\ 
{\tt\small Department of Computer Science, Union College, Schenectady, NY, USA} \\
    {\tt\small mayora@union.edu, subedi@union.edu, rieffelj@union.edu}
}

\begin{document}

\maketitle
\thispagestyle{empty}
\pagestyle{empty}

\begin{abstract}
Soft robotics leverages deformable materials to develop robots capable of navigating unstructured and dynamic environments. Silicone Voxel-Based Soft Robots (Silibots) are a type of pneumatically actuated soft robots that rely on the inflation and deflation of their voxels for shape-shifting behaviors. However, traditional pneumatic actuation methods—high pressure solenoids, medical diaphragm pumps, micro compressors, compressed fluid— pose significant challenges due to their limited efficacy, cost, complexity, or lack of precision. This work introduces a low-cost and modular syringe pump system, constructed with off-the-shelf and 3D printed parts, designed to overcome these limitations. The syringe pump system also enhances actuation with the unique ability to pull a vacuum as well pump air into the soft robot. Furthermore, the syringe pump features modular hardware and customizable software, allowing for researchers to tailor the syringe pump to their requirements or operate multiple pumps simultaneously with unique pump parameters. This flexibility makes the syringe pump an accessible and scalable tool that paves the way for broader adoption of soft robotic technologies
in research and education.

\end{abstract}

\begin{figure}[h!]
    \centering
    \includegraphics[width=0.5\textwidth]{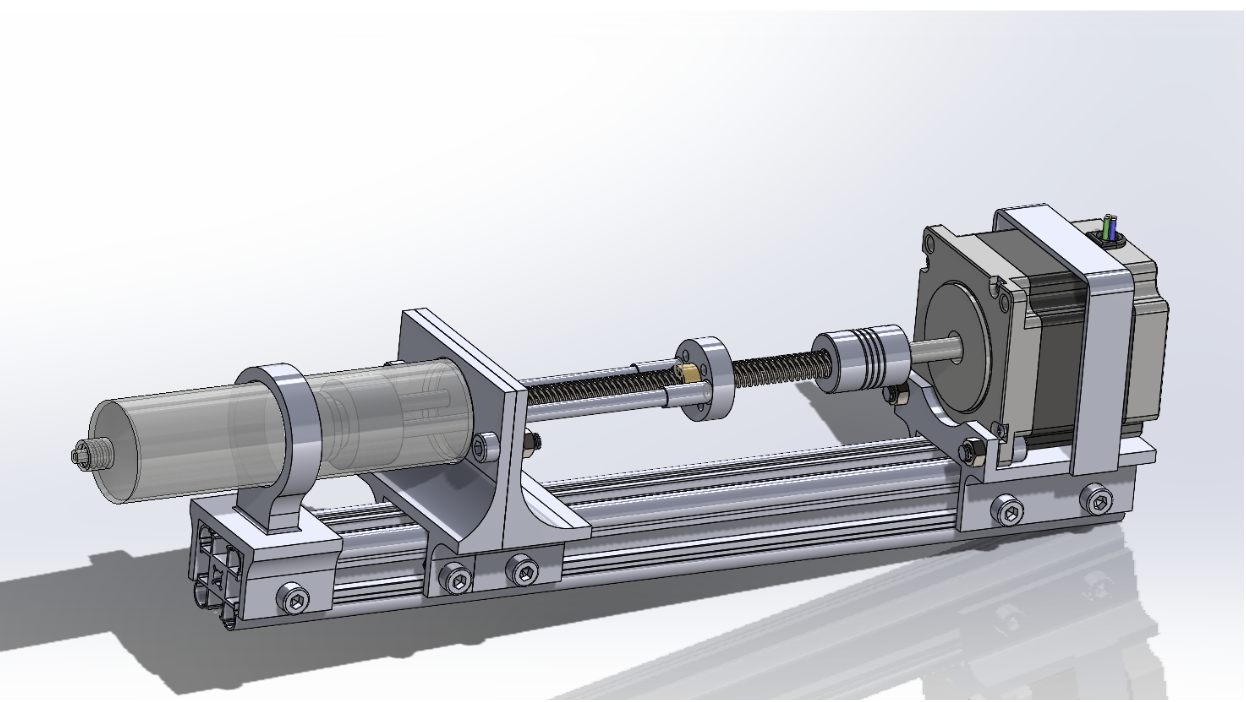}
    \caption{A 3D assembly of the syringe pump system. The components are observed to be locked onto an aluminum frame using nuts and screws.  }
    \label{fig:fitness_vs_trials}
\end{figure}

\section{Introduction}
Soft robotics aims to replicate the adaptability and compliance of biological systems enabling robots to perform tasks in unstructured and uncertain environments. Pneumatically actuated soft robots are prevalent in the soft robotics industry due to their adaptability, lightweight, low pollution, and resemblance to an organic being \cite{k1}. Silicone Voxel-Based Soft Robots (Silibots) are an example of pneumatically actuated soft robots that achieve actuation through the inflation/deflation of their select voxels. However, existing actuation mechanisms often present significant drawbacks: 
\begin{itemize}
\item \textbf{High-pressure Solenoids:} These systems lose precise control over air displacement and pose a risk of damaging the silicone voxels due to their high pressure. Not only do these systems require a pressurized system, they also cannot pull a vacuum. Pressurized systems with compressed fluids (most commonly CO2 or air) are risky due to clogs and leaks, and make the soft robot system more heavy and complex than needed while also not being sustainable for long-term use \cite{k2}.
\item \textbf{Medical-grade Syringe Pumps:} Although precise, these pumps are prohibitively expensive, costing $2,000–$5,000 per unit, making them inaccessible to many researchers.
\item \textbf{Diaphragm Pumps:} While these are one of the common ways of actuation in soft robotics, micro compressors such as this are only effective when low pressure and low flow rates are favored \cite{k2}. A large Parker Hargraves micro-compressor is rated to pump air at 11 L/minute, but the soft robot increases in complexity as different configurations and circuits of valves and compressors must be integrated \cite{k3}\cite{k4}. Extremely small micro-compressors are also commercially available but do not work effectively in actuating soft robots due to low pressure and airflow.

\end{itemize}
Moreover, other mechanisms have been applied before such as hand-held pump \cite{k5}, combustion, compressed fluid, etc. However, one main thing all the mechanisms have in common is that they cannot adequately do both, push and pull air (at high frequencies), limiting the level of actuation possible. Additionally, most of these methods have significantly high costs which make it difficult to scale up; as a result, some soft robots can be observed to have all of their actuators in phase, the same amplitudes and frequencies \cite{k5}.

To address these challenges, we developed a low-cost syringe pump system that balances precision, modularity, and affordability. This system is specifically designed to inflate and deflate silicone voxels reliably, providing researchers with a practical alternative to existing solutions. These pumps were developed to be compatible with the BROOKS simulator, where hyper-realistic simulation is achievable \cite{a1}. 

\section{Design and Development}

\begin{figure}[h!]
    \centering
    \includegraphics[width=0.5\textwidth]{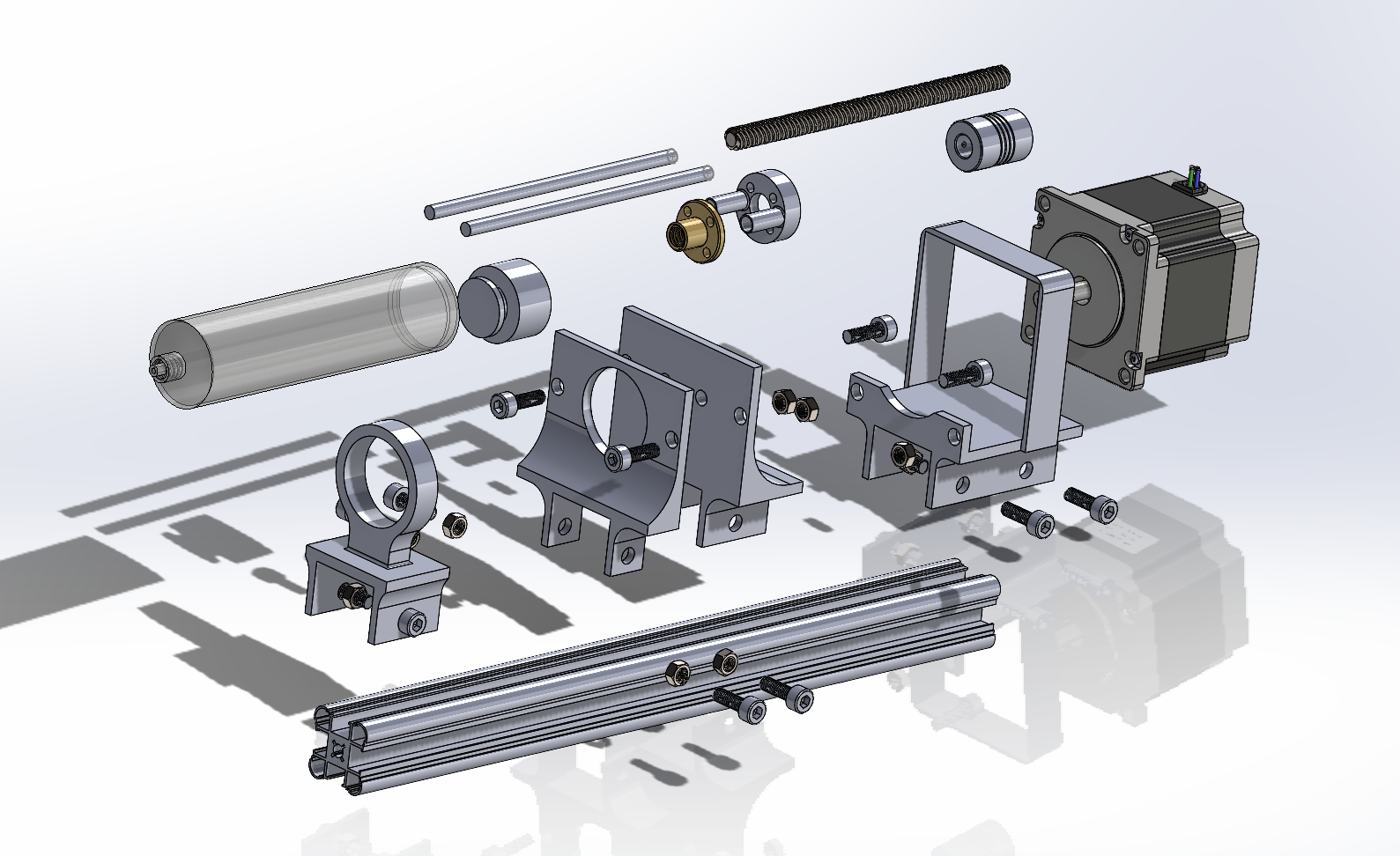}
    \caption{An exploded view of the syringe pump, showcasing all the different components that make up the system.   }
    \label{fig:fitness_vs_trials}
\end{figure}

The syringe pump was designed to meet key requirements, which include cost-effectiveness, precision, modularity, accessibility, and ease of integration into a multi-pump system. A crucial requirement was also that the syringe pump must be able to repeatedly push air in and out of the syringe pump.

As can be seen in \textbf{Figure 1}, all components that make up the syringe pump are mounted on an aluminum frame. The components are secured in place using screws and nuts, which firmly lock when tightened. This approach causes the system to be fast to construct (${\sim}$20 minutes) and modular, which makes the syringe pump easily repairable if necessary. Furthermore, all custom components have been produced via 3D printing with PLA at a 100\% fill density, providing rigidity and reliability. The use of 3D printing, along with easily accessible materials (medical syringe pump, aluminum rods, screws, nuts, etc.), also enables ease of repeatability to the pump system. \textbf{Figure 2} portrays the exploded view of all the individual components that make up the system.

Moreover, as the syringe pump is required to be reversible to pull air in, a stepper motor is utilized with a stepper motor driver that provides more precision. An Arduino microcontroller is used with customizable software that can be scaled to integrate multiple motors, where the period, amplitude, duty cycle, and other parameters can be varied. The stepper motor is used in combination with a threaded rod and a lead screw nut, which converts the rotational motion of the threaded rod into linear motion and delivers force to the syringe. Linear ball bearings are also used to maintain alignment, minimize friction, and eliminate excess vibrations. The linear ball bearings cause the system to be quieter and ensure the longevity of all other components.

\section{Applications and Benefits}

\begin{figure}[h!]
    \centering
    \includegraphics[width=0.5\textwidth]{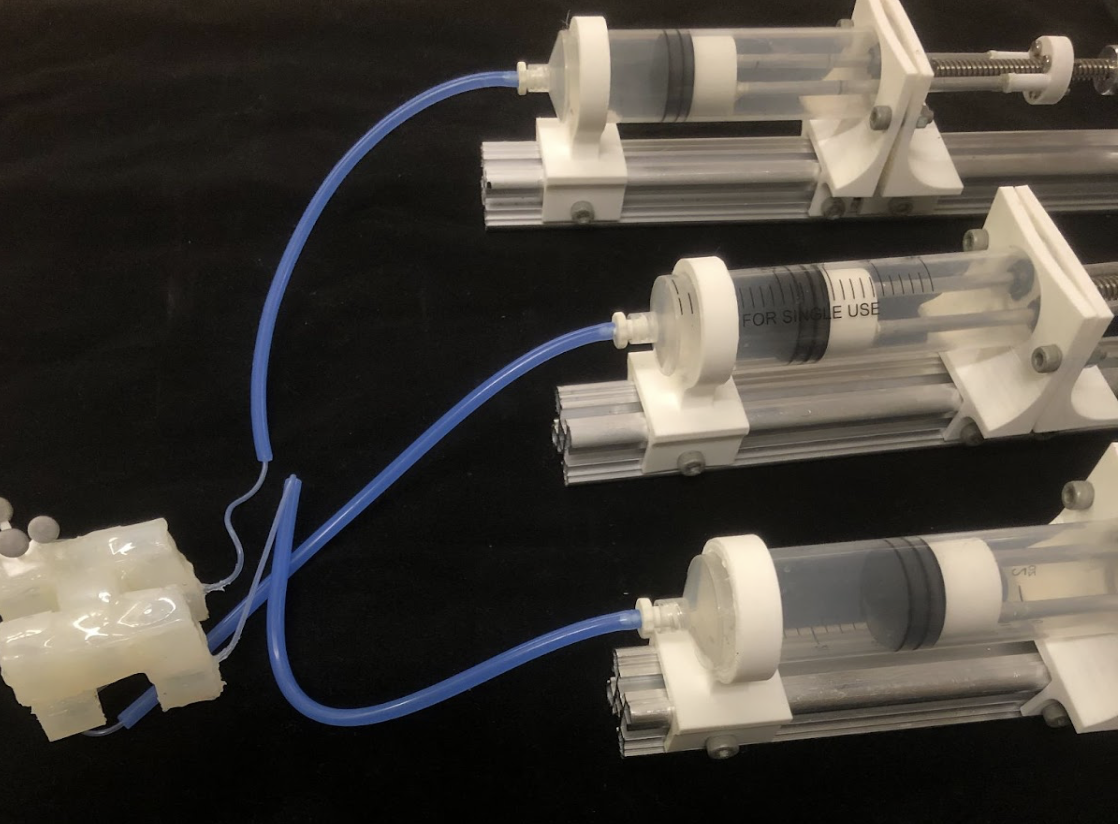}
    \caption{Three syringe pump systems running in parallel driving a Silibot, demonstrating the scalable usage of the design.}
    \label{fig:fitness_vs_trials}
\end{figure}
The syringe pump system addresses key limitations of existing actuation mechanisms:
\begin{itemize}
    \item \textbf{Accessibility:} At under \$200 per unit, pumps provide an affordable alternative to medical grade systems, significantly lowering the entry barrier for researchers and educators.
    \item \textbf{Scalability:} The modular design allows multiple pumps to be integrated into the set-up, allowing multiple pumps to run in parallel, as shown in figure 3. 
    \item \textbf{Reliability:} By ensuring precise control of air displacement, the pump system mitigates the risks associated with high-pressure systems.
\end{itemize}
Beyond its utility in Silibots, the syringe pump system has significant potential in other soft robotics applications where pneumatic actuation is required. For example, technologies such as PneUI\cite{a2} have demonstrated the transformative power of pneumatically actuated soft composite materials in creating shape-changing interfaces. The syringe pump system presented here could be used in this or similar use cases, providing an affordable and modular alternative for developing pneumatically actuated systems in human-material interaction, shape-shifting devices, and other applications.

In addition, this pump would be a suitable alternative for single-use pumps or high pressure pumps that offer less precision \cite{a3}. This can be more beneficial when it comes to research applications as more precise control is available for the pumps.

\section{Conclusion and Future Work}
This study introduces a low-cost syringe pump system that combines affordability, precision, and modularity, addressing the primary limitations of existing actuation mechanisms for soft robotics. The system's customizable software and scalable design make it a versatile tool for advancing research on silicone voxel robots such as Silibots. Future work will explore additional applications and further optimize the system for enhanced performance in diverse experimental setups.

\bibliographystyle{unsrt}

\bibliography{refs}
\end{document}